\documentclass[]{style}

\usepackage{titletoc}
\usepackage[toc,page,header]{appendix}

\usepackage{minitoc}
\usepackage{natbib}
\usepackage{CJKutf8}

\usepackage{xargs}

\usepackage{todonotes}

\usepackage{multirow}

\usepackage{cleveref}

\usepackage{amsmath}
\usepackage{dsfont}

\usepackage{svg}

\usepackage{mathrsfs}
\usepackage{adjustbox}
\usepackage{multirow}
\usepackage{multicol}
\usepackage{tcolorbox}
\usepackage{changepage}
\usepackage{enumitem}
\usepackage{graphicx}
\usepackage{amssymb}
\usepackage{xcolor}
\usepackage{float}
\usepackage{multirow}
\usepackage{threeparttable}
\usepackage{graphicx}
\usepackage{subcaption}
\usepackage{algorithm}
\usepackage{algpseudocode}
\usepackage{wrapfig}
\usepackage[table]{xcolor}  
\usepackage{colortbl}       
\usepackage{tabularx} 
\usepackage{makecell} 
\usepackage[dvipsnames]{xcolor}

\newcolumntype{g}{>{\columncolor{gray!10}}c} 

\definecolor{catgray}{gray}{0.9}
\definecolor{skyblue}{rgb}{0.53,0.81,0.92} 

\colorlet{skyblue!30}{skyblue!30!white} 

\definecolor{customblue}{RGB}{70,130,180}  

\newtcolorbox{evolbox}[2][]{%
  enhanced,
  colframe=customblue,
  colback=white,
  coltitle=white,
  rounded corners,
  boxrule=1pt,
  titlerule=0pt,
  toptitle=1mm,
  bottomtitle=1mm,
  fonttitle=\bfseries,
  width=#2\textwidth, 
  #1
}

\usepackage{url}

\PassOptionsToPackage{table,xcdraw}{xcolor}
\usepackage{placeins}
\usepackage{pifont}

\definecolor{RowBlue}{HTML}{E9F2FB}
\definecolor{RowRed}{HTML}{F9EAEA}
\definecolor{Top1}{HTML}{50DB4B} 
\definecolor{Top2}{HTML}{A5FFA2} 
\definecolor{Top3}{HTML}{D9FFD9} 
\definecolor{Sub1}{HTML}{C7DBF2}
\definecolor{Sub2}{HTML}{E4E4E4}

\renewcommand{\emph}[1]{\textit{#1}}

\usepackage{fixmath,mathtools,nicefrac,resources/mmstyle}
\usepackage{anyfontsize}
\usepackage[T1]{fontenc}
\usepackage{ulem}

\definecolor{my_green}{RGB}{51,102,0}
\definecolor{my_red}{RGB}{204, 0, 0}
\definecolor{myblue}{RGB}{218,232,252}
\definecolor{mygray}{RGB}{220,220,220}
\definecolor{mypink}{RGB}{251,49,153}

\title{InfinityEdit: Infinite Video Editing with a Lightweight Edit-Ignition Adapter}

\author[1, \circ]{Yunze Tong}
\author[1,2 \circ \dagger]{Mushui Liu}

\author[1]{Canyu Zhao}
\author[2]{Shiyi Zhang}
\author[1]{Didi Zhu}
\author[2]{Peng Zhang}
\author[2]{Wanggui He}

\author[2]{Jinlong Liu}
\author[2]{Ying Chen}

\author[2 \dagger]{Hao Jiang}
\author[2]{Pipei Huang}
\author[2]{Bo Zheng}

\makeatletter
\renewcommand\author[2][]{%
  \addtolist[#1]{#2}{\authorlist}{\authorformat}{}
  \renewcommand\author[2][]{%
    \addtolist[##1]{##2}{\authorlist}{\authorformat}{, }%
  }%
}
\makeatother

\affiliation[1]{Zhejiang University}
\affiliation[2]{Alibaba Group}
\contribution[\circ]{Equal contribution}
\contribution[\dagger]{Corresponding Author}

\abstract{
  With large pretrained models, existing methods have effectively improved instruction-based video editing.
  However, most of them rely on an \textit{in-place} editing assumption. 
  They align the edited video with the given source clip frame by frame over a fixed time span. 
  This pattern fails for open-ended streams, \textit{e.g.}, restyling a live game or applying a camera move to an ongoing shot. 
  In such cases, edits must extend to future frames as they arrive, rather than be applied to a static input clip. 
  In this paper, we study this setting and name it \textbf{infinite video editing}: given a preceding segment and an edit request, a model must generate the next segment that continues the stream while applying the requested edit. 
  This process repeats as an unbounded sequence of edit instructions arrives. 
  This task brings two challenges: the edit must be a faithful continuation rather than a frame-wise rewrite, and generation quality must remain stable as edits accumulate.
  To address them, we first design a data-collection pipeline for infinite video editing. Based on the collected data, we propose \textbf{InfinityEdit}, a lightweight edit adapter that equips a streaming video generator with unbounded editing ability.
  The adapter contains three attention modules. History cross-attention guides the denoising frames using the input frames. Temporal causal self-attention keeps temporal cues flowing only from earlier frames to later ones. Edit cross-attention injects the edit request into generation.
  During inference, the adapter is activated only in the chunk where an edit request arrives. Subsequent chunks are generated by the original model with a reset anchor frame. This scheme applies the edit while preserving the original model's infinite generation ability.
  Extensive experiments show that InfinityEdit faithfully continues the stream under each edit, and stays stable over unbounded edit sequences.
}

\checkdata[
\raisebox{-0.25em}{\includegraphics[height=1.1em]{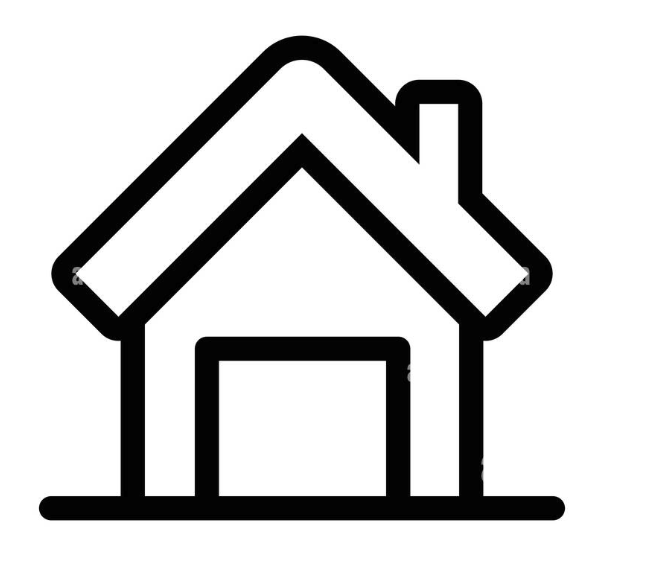}}~~Project Page]{\href{https://yunzetong.github.io/InfinityEdit}{\texttt{https://yunzetong.github.io/InfinityEdit}}
\\[-1.5ex]}

\checkdata[
\raisebox{-0.25em}{\includegraphics[height=1.1em]{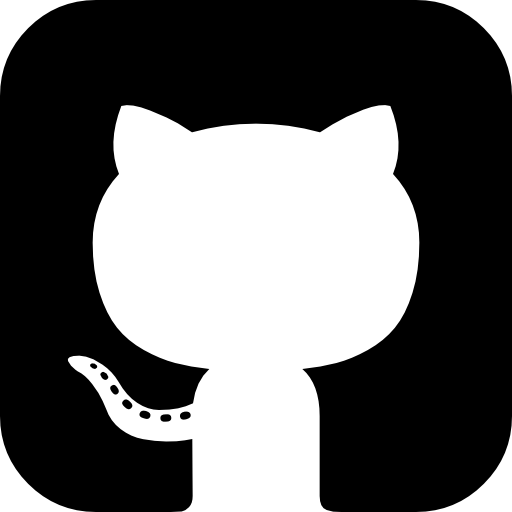}}~~Github]{\href{https://github.com/YunzeTong/InfinityEdit}{\texttt{https://github.com/YunzeTong/InfinityEdit}} 
\\[-1.5ex]}

\begin{document}
\maketitle

\section{Introduction}
Diffusion transformers can generate high-fidelity videos from text prompts~\cite{CogVideoX, Wan, HunyuanVideo, Seedance2.0, Sora2}. 
This progress has also improved instruction-based video editing, where a model modifies a given clip according to a natural-language edit request~\cite{InsViE, Ditto, OpenVE, VAP}. 
However, existing editors are built for fixed temporal windows. 
They follow an \textit{in-place} editing setting. 
The output has the same duration as the source clip, and is aligned to the same temporal span. 
This makes editing a rewrite of a finite clip. 
It does not define how an edit should continue beyond that clip, where future frames have not yet appeared. 
This fixed-window setting is not suitable for open-ended video streams.
In applications such as restyling a live game stream or applying a camera move to an ongoing shot, new content keeps arriving. 
The edit must therefore extend to future frames, rather than be applied once to a fixed clip. 
We study this setting and name it \textbf{infinite video editing}. 
Given a preceding video segment and an edit request, a model generates the next segment that continues the stream while applying the edit. 
This process repeats as an unbounded sequence of edit requests arrives.

Beyond faithfully applying the requested edit, infinite video editing brings two additional challenges. 
First, the edited segment must be a faithful continuation rather than a frame-wise rewrite. 
It lies beyond the input window, so the model cannot treat editing as a timestamp-wise transformation of the input clip. 
At the same time, entities unaffected by the edit should remain consistent with the input. 
Second, generation quality must remain stable as edits accumulate. 
Each generated segment becomes the history for the next one, so errors can propagate over time. 
To meet these requirements, a natural attempt is to reuse a streaming generator and switch its scene prompt when an edit request arrives. 
However, such generators are trained to continue the current video from its history, not to apply a specified relational edits. 
Their stability mechanisms can also conflict with the edit. 
For example, anchor frames~\cite{zhang2025framepackv1, yang2026anchor} and cached multi-scale history~\cite{longlive, helios} tend to pull later segments back toward the unedited content. 
The edit can therefore be weakened or ignored. 
This makes a dedicated design necessary for infinite video editing, rather than a simple prompt switch.

To address these challenges, we first build a data-collection pipeline that synthesizes \texttt{(source, edit, target)} triplets for this task. 
Based on this data, we propose \textbf{InfinityEdit}, a lightweight edit adapter that equips a streaming video generator with infinite editing ability. 
We adopt Helios~\cite{helios} as the backbone and keep it frozen. This preserves its prior for long, stable video, while only the edit adapter is trained. 
Our adapter is designed with three attention modules. 
First, history cross-attention conditions the new denoising chunk on the provided input frames. 
Second, temporal causal self-attention keeps temporal cues flowing only from earlier frames to later ones within the chunk. 
Third, edit cross-attention effectively injects the edit instruction into the denoising process. 
To improve stability across multiple rounds of generation, we train the adapter with history corruption, so it can tolerate imperfect generated frames as input. 
We also adopt a two-phase curriculum: the model first learns to apply the edit and then focuses on refining details. 
At inference, we uses an "ignite-then-continue" strategy.
The adapter is activated only on the chunk where an edit instruction arrives, which ignites the edit in the current generated chunk. 
The frozen generator then continues generation and carries the edit forward until next edit instruction arrives. 
An anchored sliding-window history bounds memory and re-anchors generation to the edited content. 
In this way, InfinityEdit uses the adapter for editing while preserving the long-video prior of the frozen generator.
These designs together support an infinite sequence of edits. 
Experiments on out-of-distribution cases show that InfinityEdit continues the stream under each edit and remains stable over long edit sequences.

Our contributions are summarized as follows:
\begin{itemize}
    \item We study \textbf{infinite video editing}, a task where edits are repeatedly applied to an ongoing video stream. 
    Beyond edit alignment, we identify two key challenges: faithful continuation and stability under repeated editing.
    \item We design a data-collection pipeline for this task. It synthesizes data with edit-type-aware generation and provides supervision for model training.
    \item We propose \textbf{InfinityEdit}, a lightweight edit adapter that gives a streaming video generator infinite editing ability. The adapter applies each edit when it arrives, and the frozen generator then carries the edited content forward across later chunks.
    \item Extensive experiments show that InfinityEdit could effectively handle the infinite video editing task.
\end{itemize}

\section{Related Work}\label{sec:related-work}

\textbf{Video Generation and Editing}

The success of text-to-image diffusion models~\cite{score-based-method,stable-diffusion,flux,sd35, liu2025tfcustom, liu2024llm4gen, zhao2026marble, tong2025decoding, tong2026alleviating} has encouraged extending diffusion models to video. Diffusion Transformers (DiTs)~\cite{Peebles2022DiT} are now widely used as video backbones, as their spatio-temporal attention can model motion across frames. This paradigm has given rise to a family of open foundational generators, including CogVideoX~\cite{CogVideoX}, Wan~\cite{Wan}, and HunyuanVideo~\cite{HunyuanVideo, HunyuanVideo1.5}. Newer models, such as SkyReels-V4~\cite{SkyReels-V4}, Sora~2~\cite{Sora2}, and Seedance~2.0~\cite{Seedance2.0}, further improve visual quality, motion realism, and video length.
Beyond video synthesis, another line of work studies in-place video editing. Instruction- and reference-driven methods, such as InsViE~\cite{InsViE}, Ditto~\cite{Ditto}, OpenVE~\cite{OpenVE}, and Kiwi-Edit~\cite{Kiwi-Edit}, let users specify edits with natural-language commands or visual examples. Other methods focus on semantic control and effect transfer, such as RefVFX~\cite{jones2026tuning} and Video-As-Prompt~\cite{VAP}. These methods edit a given source clip within the same temporal window, making them well suited to in-place editing applications.

\textbf{Long Video Generation and Streaming Video Generation}

Long video generation is often formulated as autoregressive continuation. Each new chunk is denoised conditioned on a window of previously generated frames. CausVid~\cite{causvid}, for example, distills a bidirectional diffusion teacher into a causal autoregressive student. A key challenge in this setting is drift. The model is trained with clean ground-truth history, but at inference it conditions on its own imperfect outputs. To reduce this exposure bias, prior works try several strategies: history-frame corruption during training~\cite{chen2025diffusion}, self-rollout with train-as-infer objectives~\cite{huang2025selfforcing}, error banks that recycle prediction errors~\cite{li2025stable}, and rolling-window denoising over progressively noised frames~\cite{liu2025rolling}.
A second line of work focuses on low-latency streaming video generation. In this setting, each chunk must be sampled immediately. The main costs come from per-chunk sampling and history conditioning. Step distillation reduces the sampling cost by shortening denoising to a few steps, with methods such as reward-guided distillation~\cite{lu2025reward}, autoregressive diffusion distillation~\cite{zhu2026causal, zhao2026causal}, and flow-map distillation~\cite{gu2026anyflow}. For history conditioning, sparse context retrieval~\cite{cai2026moc}, context compression~\cite{zhang2025framepackv1}, and hierarchical history caching~\cite{longlive, longlive_2.0} keep attention windows and KV caches bounded. These techniques support streaming systems and interactive applications, including StreamDiffusionV2~\cite{feng2025streamdiffusionv2}, real-time avatars~\cite{huang2025liveavatarstreamingrealtime, sun2026streamavatar}, and humanoid video generation~\cite{zhao2024moviedreamer, wang2026flowact}.

\textbf{Streaming Editing and Multi-shot Video Generation}

Building on the above progress in video editing and streaming generation, recent work has explored two related settings. The first is streaming video editing, which edits a live stream frame by frame with strict content preservation. SANA-Streaming~\cite{zhao2026sana} and LiveEdit~\cite{liveedit} further target real-time, low-latency editing on consumer hardware. In essence, these systems perform in-place editing on streaming inputs. The second is multi-shot long-video generation. In this setting, a generator continues a stream while the conditioning prompt is changed to introduce new content. LongLive~\cite{longlive}, Anchor Forcing~\cite{yang2026anchor}, and CausalCine~\cite{causalcine} follow this setting. This line is generation rather than editing. Each new prompt describes the target scene directly, rather than an edit to apply to existing content. 
In this paper, our task is related to both settings but differs in its target: it applies edit requests to an ongoing stream and requires the edited result to continue into future segments. We define this setting in Section~\ref{subsec: task description preliminary}.

\section{Preliminaries}

\subsection{Infinite Video Editing Task}\label{subsec: task description preliminary}

We define infinite video editing by the temporal relation between input and output. Traditional \textbf{in-place} video editing operates on a fixed time span. Given a source clip $V_\text{src} \in \mathbb{R}^{T \times C \times H \times W}$ and an edit instruction, it produces an edited clip $V_\text{edit} \in \mathbb{R}^{T \times C \times H \times W}$. The output has the same duration as the source clip and is aligned to the same temporal span $T$. Infinite video editing instead produces a temporally \textbf{subsequent} segment. Given a preceding segment $V_\text{pcd} \in \mathbb{R}^{T_1 \times C \times H \times W}$ and an edit instruction $c_\text{edit}$, we seek a generator
\begin{equation}
V_\text{tgt} = \mathcal{G}(V_\text{pcd}, c_\text{edit}), \qquad V_\text{tgt} \in \mathbb{R}^{T_2 \times C \times H \times W},
\end{equation}
where $V_\text{tgt}$ continues $V_\text{pcd}$ in time while satisfying $c_\text{edit}$. In general, $T_1 \neq T_2$, and target frames do not directly correspond to any input frame. We further consider the infinite regime, where a stream is edited by a sequence of edit instructions $\{c_\text{edit}^{(i)}\}_{i \geq 1}$:
\begin{equation}
V^{(i)} = \mathcal{G}\big(V^{(i-1)}[-T_1{:}],\, c_\text{edit}^{(i)}\big), \qquad V^{(0)} = V_\text{init},
\end{equation}
where $V^{(i-1)}[-T_1{:}]$ denotes the most recent $T_1$ frames generated so far. Each edited segment then becomes the conditioning history for the next.

This task brings three challenges. The first is edit alignment: $V^{(i)}$ should faithfully realize $c_\text{edit}^{(i)}$, as in any instruction-based editing task. The second is faithful continuation. Since the target segment has no directly aligned source frames, the model cannot edit by simply transforming existing frames. It must generate a natural continuation from the input history. What should be preserved across the boundary depends on the edit type. For example, style transfer should change the visual appearance while preserving object motion and camera movement. A camera-move edit should instead preserve the subject and scene appearance while changing the viewpoint. The third is stability under repeated editing. Each generated segment becomes the conditioning history for the next, so errors can be fed back over time. Quality must remain stable as edits accumulate. The latter two challenges are specific to infinite video editing and are absent from the in-place setting. This task has practical value for on-the-fly editing of open-ended streams, such as continuously restyling game footage or applying camera moves to an ongoing shot.

\subsection{Base Model: Helios}\label{subsec: model preliminary}

Helios~\cite{helios} is a 14B autoregressive video diffusion transformer for real-time long-video generation. Helios-Distilled is its few-step distilled variant. Instead of generating a fixed-length clip in a single pass, Helios generates video chunk by chunk. It concatenates the clean history with the noisy target as input and denoises only the target. The history is kept clean and serves as the conditioning context. Each generated chunk is then appended to the history, allowing generation to continue. To keep the cost bounded as the video grows, Helios compresses the history into a hierarchical multi-scale memory with a constant token budget. Its distilled variant supports few-step, real-time inference at the 14B scale. The model is conditioned on the initial input, without an explicit design for injecting new edit instructions during intermediate generation. In this paper, we adopt Helios-Distilled as the backbone of our infinite video editor, which will be detailed in Section~\ref{subsubsection: base model}.

\section{Data Collection Pipeline}

\begin{figure}[t]
    \centering
    \includegraphics[width=\linewidth]{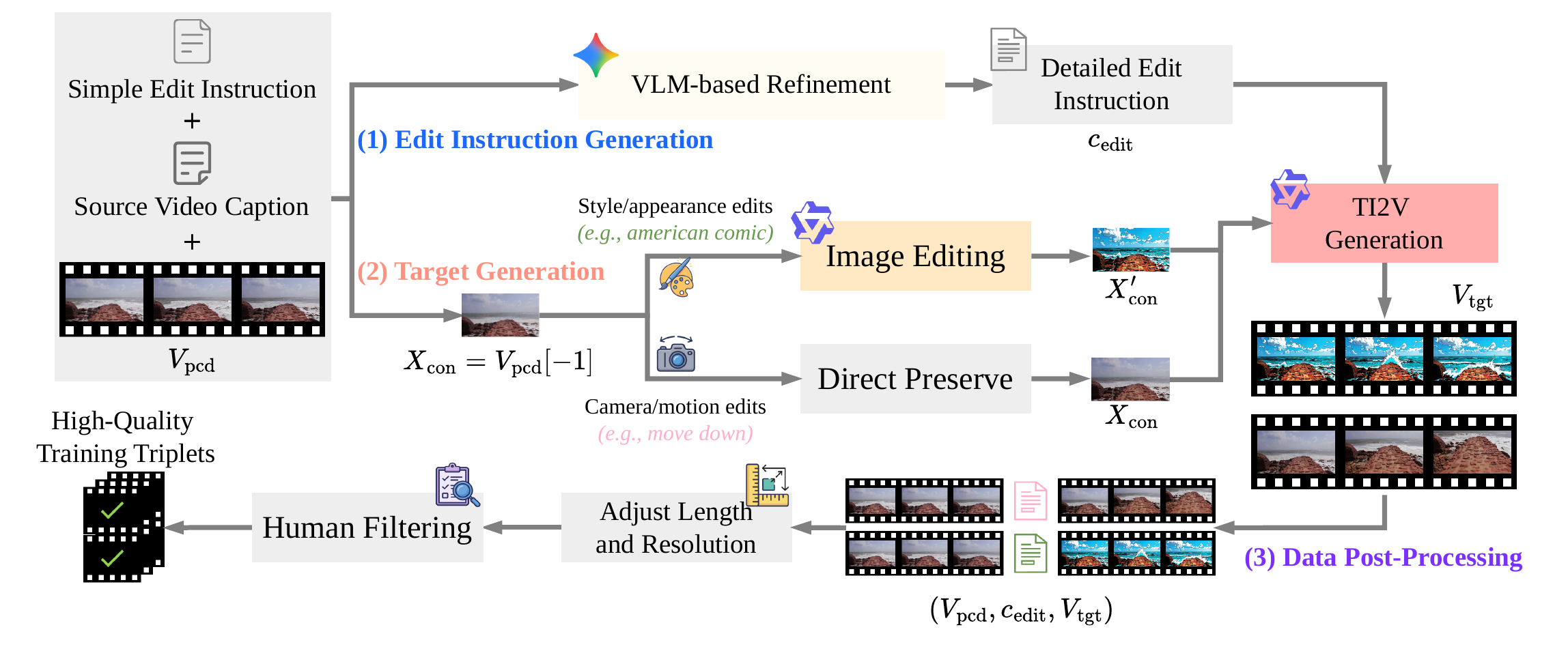}
    \caption{Our data collection pipeline.}
    \label{fig: data collection pipeline}
\end{figure}

As described in Section~\ref{subsec: task description preliminary}, training a model for our task requires triplets $(V_\text{pcd}, c_\text{edit}, V_\text{tgt})$. $V_\text{pcd}$ is the preceding video. $c_\text{edit}$ is the edit instruction. The target video $V_\text{tgt}$ continues from $V_\text{pcd}$ while applying $c_\text{edit}$.
We construct the three parts in separate steps. Source videos $V_\text{pcd}$ are sampled from UltraVideo~\cite{ultravideo} dataset and resized to a fixed resolution and frame count. Basic edit types are drawn from representative cases used by VAP~\cite{VAP}. They are then used to construct concrete edit instructions $c_\text{edit}$. Target videos $V_\text{tgt}$ are synthesized with the image-to-video model Wan2.2-I2V-A14B~\cite{wan2025}. The following paragraphs describe edit instruction generation, edit-type-aware target video generation, and data post-processing. The full data generation pipeline is shown in Figure~\ref{fig: data collection pipeline}.

\paragraph{Edit Instruction Generation.}
Our basic edit types are sampled from the instruction sets used in VAP~\cite{VAP}. These edit types are abstract phrases and are not tied to a specific source video. Thus, they cannot be used directly as edit instructions. To generate usable $c_\text{edit}$, we first group source videos $V_\text{pcd}$ by the main entities in their captions. We then pair each group with suitable edit types. Given the source caption and a simple edit type, we use Gemini 3 Flash~\cite{team2023gemini} to expand it into a detailed edit instruction $c_\text{edit}$. In this case, the generated instruction effectively describes both the edit behavior and the specific change to apply.

\paragraph{Edit-type-aware Target Video Generation.} 
Then we begin to synthesize target videos $V_\text{tgt}$ from $V_\text{pcd}$ and $c_\text{edit}$. Since $V_\text{tgt}$ should continue from $V_\text{pcd}$, its first frame should match the end of the source video. We take the last source frame as the connection frame, $X_\text{con}=V_\text{pcd}[-1]$, and use it to start target generation. Different edit types require different modification of this frame. A style change should show the new appearance immediately, while a camera-motion edit should keep the scene consistent at the boundary and only alter the viewpoint. We therefore group edit types into four main categories and process $X_\text{con}$ according to the group. For appearance edits such as style changes, we first use Qwen-Image-Edit-2511~\cite{wu2025qwenimagetechnicalreport} to obtain an edited connection frame $X'_\text{con}=\text{Edit}(X_\text{con}, c')$, where $c'$ describes the intended change. We then use $X'_\text{con}$ as the first frame of $V_\text{tgt}$. For edits that mainly change the camera view, we keep $X_\text{con}$ unchanged and use it directly as the first frame. 
We then obtain $V_\text{tgt}$ through image-to-video generation with Wan2.2-I2V-A14B.

\paragraph{Data Post-processing}

After collecting the triplets $(V_\text{pcd}, c_\text{edit}, V_\text{tgt})$, we first adjust them to match the input format of our backbone. As introduced in Section~\ref{subsec: model preliminary}, our base model is Helios-Distilled. We therefore resize all videos to its default training resolution. We also clip the frame counts so that the ratio $\frac{T_1}{T_2}$ matches the lengths of the full input window and the output window of the model. 
We then filter the triplets by manual scoring along four criteria: the alignment between $V _\text{tgt}$ and $c _\text{edit}$, the consistency between $V _\text{pcd}$ and $V _\text{tgt}$, the rationality of $V _\text{tgt}$, and the visual quality of $V _\text{pcd}$. 
Alignment measures how well the target video fulfills the instruction, \textit{i.e.}, whether $V _\text{tgt}$ conveys the intended edit type and realizes the other semantics described in $c _\text{edit}$. Consistency measures how well $V _\text{tgt}$ continues from $V _\text{pcd}$, based on whether the main subjects remain stable and the scene agrees with the final source frame. Rationality checks whether the generated video contains implausible artifacts, such as a person with extra hands or an unnatural change in a subject's appearance. Visual quality measures the quality of the selected source clip from UltraVideo. Each criterion is rated on a scale from 1 to 4, with 4 as the best and 1 as the worst, and the scoring is carried out by 20 human labelers.

\section{Methodology}\label{sec:method}

To achieve infinite video editing, we attach a lightweight edit adapter to a frozen streaming video generator. The system also includes a training recipe and an inference pipeline for repeated editing. Section~\ref{sec:method:arch} describes the architecture, including the frozen backbone and the adapter block that injects edit instructions while preserving temporal consistency. Section~\ref{sec:method:train} describes how we train the adapter to acquire editing capability. Section~\ref{sec:method:infer} describes how one trained adapter is used for infinite editing at inference.

\subsection{Model Architecture}\label{sec:method:arch}

\subsubsection{Backbone Interface and Notation}\label{subsubsection: base model}

Our editor builds on the Helios backbone described in Section~\ref{subsec: model preliminary}. Here we restate only the parts used by the adapter and define the notation for this section. The backbone generates one latent video chunk at a time. At noise level $\sigma$, a diffusion transformer denoises a chunk of $L$ frames. It is conditioned on a window of provided frames and an initial scene prompt. The history window has three temporal scales: long, middle, and short. It also starts with a single anchor frame $x_0$, which stabilizes long-range appearance. History tokens are kept at a clean noise level, so the transformer can separate the known past from the chunk being denoised. Within each transformer layer, the hidden states concatenate the history tokens and current-chunk tokens as $H=[H_{\text{hist}};\,H_{\text{cur}}]$. These per-layer hidden states, together with the multi-scale history and noise level $\sigma$, are the interfaces used by our edit adapter.

\subsubsection{Edit-Ignition Adapter}

The backbone already provides the generation prior needed for high-quality frames. Our goal is to add edit control without weakening this prior. Full fine-tuning is costly and may disturb the backbone. Prompt-only conditioning is also limited, since the edit instruction has no explicit path into the intermediate features. We therefore freeze all base weights and train small adapter blocks that operate on the backbone features. 
We insert one adapter block after each transformer layer. These blocks are the only trainable components, together with a shared module that encodes the noise level $\sigma$. Given the per-layer hidden states $H=[H_{\text{hist}};\,H_{\text{cur}}]$, an adapter block keeps the history tokens unchanged and refines the current-chunk tokens in three stages:
\begin{align}
H_{\text{cur}}^{1:L_a} &\leftarrow H_{\text{cur}}^{1:L_a} + m_\sigma\big(\textsc{HistCA}(H_{\text{cur}}^{1:L_a},\, H_{\text{hist}})\big), \\
H_{\text{cur}} &\leftarrow H_{\text{cur}} + \textsc{TempSA}(H_{\text{cur}}), \\
H_{\text{cur}} &\leftarrow H_{\text{cur}} + m_\sigma\big(\textsc{EditCA}(H_{\text{cur}},\, c_{\text{edit}})\big),
\end{align}
where the first stage updates only the leading $L_a$ frames $H_{\text{cur}}^{1:L_a}$. 
The modulation $m_\sigma(\cdot)=(1+s_\sigma)\odot(\cdot)+b_\sigma$ follows AdaLN, with scale and shift $(s_\sigma,b_\sigma)$ predicted from $\sigma$ by the shared sigma-embedding module. The output projection of each stage is zero-initialized. Thus, every adapter block starts as an identity map, and editing is learned as a residual update to the frozen prior. This keeps the original generation behavior intact. We then describe the three attention modules in the adapter below.

\begin{figure}[t]
    \centering
    \includegraphics[width=.95\linewidth]{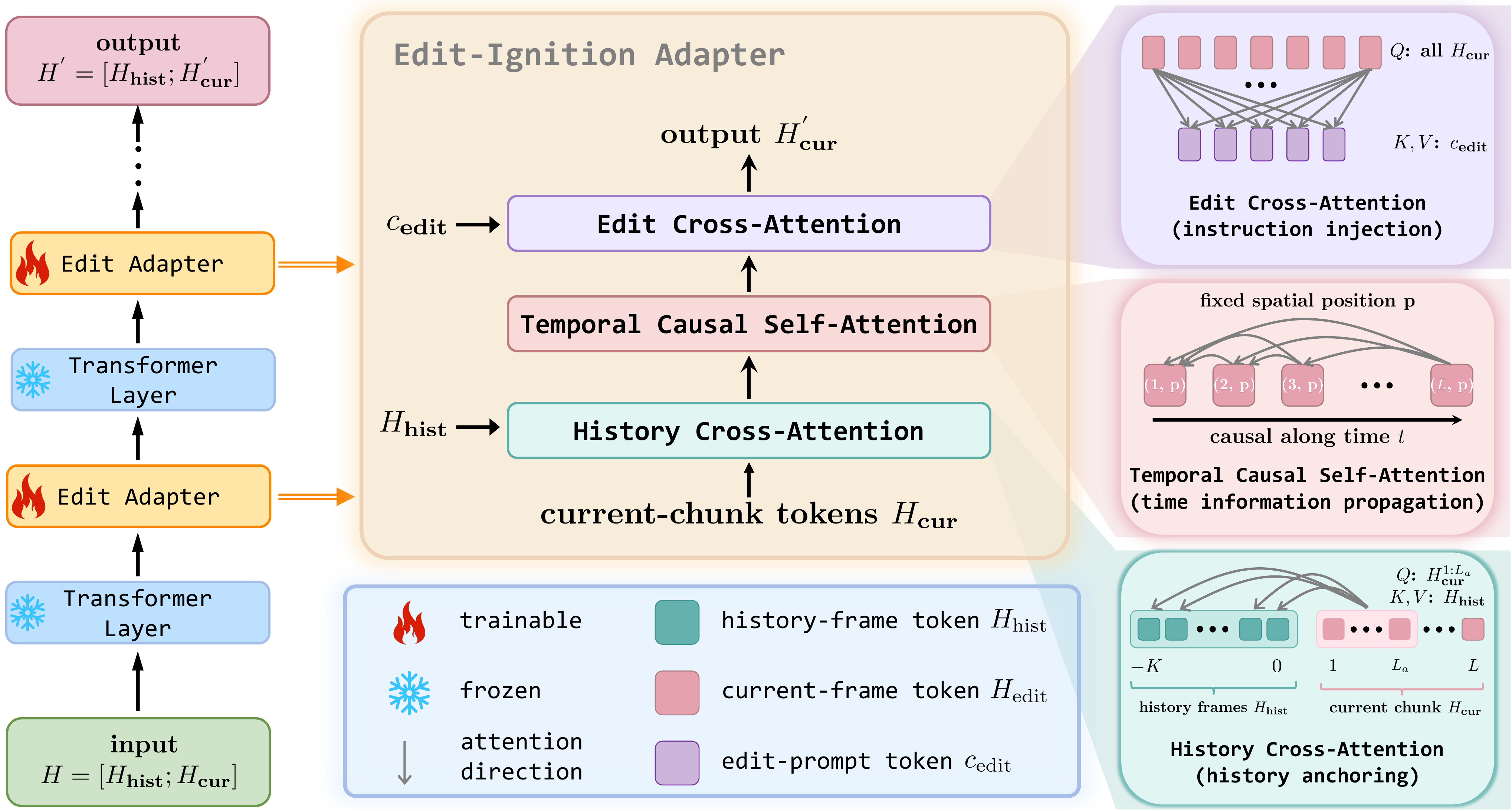}
    \caption{Our adapter's architecture. It is composed of three attention modules.}
    \label{fig: model architecture}
\end{figure}

\paragraph{History Cross-Attention for History Anchoring.}
The first stage anchors the denoising chunk to the provided history. We formulate a single attention head as $\textsc{Attn}(Q,K,V;M)=\mathrm{softmax}\big(QK^\top/\sqrt{d}+M\big)\,V$, where $d$ is the per-head key dimension. The mask is omitted ($M{=}0$) when attention is deployed. Here the queries come only from the $L_a$ leading frames, while the keys and values come from the backbone-processed history tokens:
\begin{equation}
Q=H_{\text{cur}}^{1:L_a}W_Q^{\mathrm{h}},\quad K=H_{\text{hist}}W_K^{\mathrm{h}},\quad V=H_{\text{hist}}W_V^{\mathrm{h}},\qquad \textsc{HistCA}=\textsc{Attn}(Q,K,V;0)\,W_O^{\mathrm{h}}.
\end{equation}
The result is added back to the leading frames. Restricting the queries to the first $L_a$ frames keeps this attention inexpensive. It also lets these frames take in cues from the history, which propagates to the rest of the chunk in the next stage.

\paragraph{Temporal Causal Self-Attention for Forward Propagation.}
The second stage propagates the anchored signal forward along the temporal axis. We index the current tokens by frame $t\in\{1,\dots,L\}$ and spatial position $p\in\{1,\dots,P\}$, and denote the token at $(t,p)$ by $(H_{\text{cur}})_{t,p}$. For each fixed spatial position $p$, we apply self-attention over its length-$L$ temporal sequence:
\begin{equation}
\begin{aligned}
Q = (H_{\text{cur}})_{:,p}W_Q^{\mathrm{t}},\quad K &= (H_{\text{cur}})_{:,p}W_K^{\mathrm{t}},\quad V = (H_{\text{cur}})_{:,p}W_V^{\mathrm{t}}, \\ 
\textsc{TempSA}(H_{\text{cur}})_{:,p} &= \textsc{Attn}(Q,K,V;M_{\mathrm{causal}})\,W_O^{\mathrm{t}},    
\end{aligned}
\end{equation}
where $Q=(Q_1,\dots,Q_L)$, with the same notation for $K$ and $V$. The causal mask sets $(M_{\mathrm{causal}})_{t,t'}=0$ for $t'\le t$ and $-\infty$ otherwise. We also apply a one-dimensional temporal rotary position encoding to $Q$ and $K$, separate from the backbone's spatial encoding. This causal direction matches the autoregressive order of streaming generation and lets temporal cues flow only from earlier frames to later frames.

\paragraph{Edit Cross-Attention for Instruction Injection.}
The third stage provides the path for the edit instruction to enter generation. Each token from the current denoising chunk queries $c_{\text{edit}}$:
\begin{equation}
Q=H_{\text{cur}}W_Q^{\mathrm{e}},\quad K=c_{\text{edit}}W_K^{\mathrm{e}},\quad V=c_{\text{edit}}W_V^{\mathrm{e}},\qquad \textsc{EditCA}=\textsc{Attn}(Q,K,V;0)\,W_O^{\mathrm{e}}.
\end{equation}
The output is added back to all denoising tokens, so $c_{\text{edit}}$ can affect the whole chunk. This stage provides edit control, while the preceding two stages keep the edited chunk tied to the past and coherent along time.

\subsection{Training Recipe}\label{sec:method:train}

\subsubsection{Flow-Matching Objective}

We train the adapter with flow-matching objective while keeping all base weights frozen. Given a clean target chunk $z_0$ and noise $\epsilon\sim\mathcal{N}(0,I)$, we form the interpolated state $z_\sigma=(1-\sigma)\,z_0+\sigma\,\epsilon$ and regress the velocity $v=\epsilon-z_0$:
\begin{equation}\label{eq:flow_matching}
\mathcal{L}=\mathbb{E}_{z_0,\epsilon,\sigma}\big[w(\sigma)\,\lVert v_\theta(z_\sigma,\sigma,c)-v\rVert_2^2\big],
\end{equation}
where $v_\theta$ is the adapter-augmented transformer, $c$ includes the scene prompt, edit instruction, and history condition, and $w(\sigma)$ is a sigma-dependent loss weight.

\subsubsection{History Corruption for Exposure-Bias Mitigation}

At inference, the adapter conditions on previously generated frames, which may contain errors. Motivated by prior work~\cite{chen2025diffusion, helios}, we use history corruption during training to expose the adapter to such imperfect feedback. 
With probability $p_\text{corrupt}$, each clean history latent $x$ is replaced by $\tilde{x}=\sigma_c\,\epsilon+(1-\sigma_c)\,x$, where $\epsilon\sim\mathcal{N}(0,I)$ and $\sigma_c$ is sampled independently per frame from a bounded range. With probability $1-p_\text{corrupt}$, the history remains clean. We apply corruption independently to the long, middle, and short history scales. This design helps our adapter handle imperfect history while still working with clean history.

\subsubsection{Mixture-Gaussian Sampling of Noise Levels}

We next specify how the noise level $\sigma$ is sampled during adapter training. Our backbone model uses a pyramid denoising schedule for efficient video synthesis~\cite{jin2025pyramidal, helios}. It assigns different $\sigma$ ranges to different resolution stages: high-noise steps form the low-resolution layout, and lower-noise steps refine high-resolution details. This design works well for generation, where coarse-to-fine synthesis is natural. 
For editing, however, the required change is not tied to a fixed visual scale. An edit instruction may affect viewpoint, layout, appearance, or fine details. If noise ranges are coupled with resolution stages, a lightweight adapter must learn edit control across both noise levels and spatial scales. This adds extra variation and makes the training target less focused. We therefore use a single-stage fixed-step Euler schedule for editing. This keeps the representation space consistent and lets the adapter focus on how edits act across the sampled $\sigma$ values. 
Concretely, we replace the backbone's logit-normal $\sigma$ prior with an $N$-component Gaussian mixture:
\begin{equation}\label{eq:mixture_sigma}
\sigma\sim\sum_{k=1}^{N}\pi_k\,\mathcal{N}(\mu_k,\,\delta_k^2),\qquad \sum_{k=1}^{N}\pi_k=1,
\end{equation}
where each center $\mu_k$ is one discrete $\sigma$ sampled when applying an edit at inference. The last center $\mu_N\!\approx\!0$ corresponds to the near-zero final step. Here $\delta_k$ is a small per-component standard deviation, and $\pi_k$ controls how often each center is sampled. This sampling puts more training examples near the $\sigma$ values used by the adapter during inference. We set $\pi_k$ separately for different training stages, as detailed in Section~\ref{subsubsection: two phase curriculum}.

\subsubsection{Two-Phase Curriculum: From Uniform Coverage to Detail Refinement}\label{subsubsection: two phase curriculum}

We split training into two phases. The first phase learns the basic editing ability, and the second phase improves the final editing quality. Both phases use the flow-matching objective in Eq.~\eqref{eq:flow_matching} and the mixture-Gaussian sampler in Eq.~\eqref{eq:mixture_sigma}. They mainly differ in two components: the mixture weights $\{\pi_k\}$ and the per-frame loss weight $\omega_t$. The weights $\{\pi_k\}$ decide which noise levels are sampled more often during training. The weight $\omega_t$ scales the loss for each frame in a chunk.

In the first phase, we use balanced supervision. The mixture weights are uniform ($\pi_k\!=\!1/N$), so all used $\sigma$ values during inference are covered equally. The frame weight is flat ($\omega_t\!=\!1$), so no specific frame in the generated chunk is emphasized. In the second phase, we initialize the model from the first phase and turn to refine details. We shift $\{\pi_k\}$ toward middle- and low-$\sigma$ centers, which have stronger influence on final visual details. The density of sampled $\sigma$ can be viewed in Appendix.
We also increase $\omega_t$ with the frame index, so later frames receive larger loss weights. These frames are farther from the history anchor and are more likely to lose quality. This curriculum first gives the adapter broad denoising coverage and then focuses training on later denoising steps and harder frames. Detailed configurations are reported in the Appendix.

\subsection{Inference Pipeline}\label{sec:method:infer}

\subsubsection{First-Chunk Ignition and Backbone Continuation}

For each edit round at inference, our goal is to apply the edit once and then continue the edited stream. Therefore, the adapter is used only to "ignite" the edit on the first chunk after an edit instruction arrives. Later chunks are generated by the frozen backbone with the adapter disabled. The full inference pipeline is shown in Figure~\ref{fig: method pipeline}. Concretely, the first chunk is denoised by the adapter with the single-stage Euler schedule. The following chunks are denoised by the backbone with its pyramid scheduler. This switch works because the backbone conditions each new chunk on generated history. Once the first edited chunk enters the history window, later chunks inherit the edit from that history. Thus, the instruction-conditioned path is used only where the edit is introduced, while the backbone carries the edited content forward.

\subsubsection{Low-Noise Sampling for Fine-Detail Completion}

Lower noise levels correspond to later denoising steps, where fine details are mainly formed. Some edit instructions may target these details rather than only larger appearance changes. To ensure that the ignition chunk in each edit round present such edits clearly, we add one denoising step at a near-zero $\sigma$. At this stage, the latent is close to the clean sample. The adapter can then refine the almost-clean chunk and complete fine details. We include the same low-noise step during training through the $\mu_N\!\approx\!0$ component in Eq.~\eqref{eq:mixture_sigma}. This keeps the training and inference $\sigma$ distributions aligned. The extra step improves detail quality at little cost and avoids under-training this final step.

\subsubsection{History-Window Sliding and Anchor Resetting for Infinite Editing}\label{subsubsec: history window and anchor resetting}

\begin{figure}[!t]
    \centering
    \includegraphics[width=\linewidth]{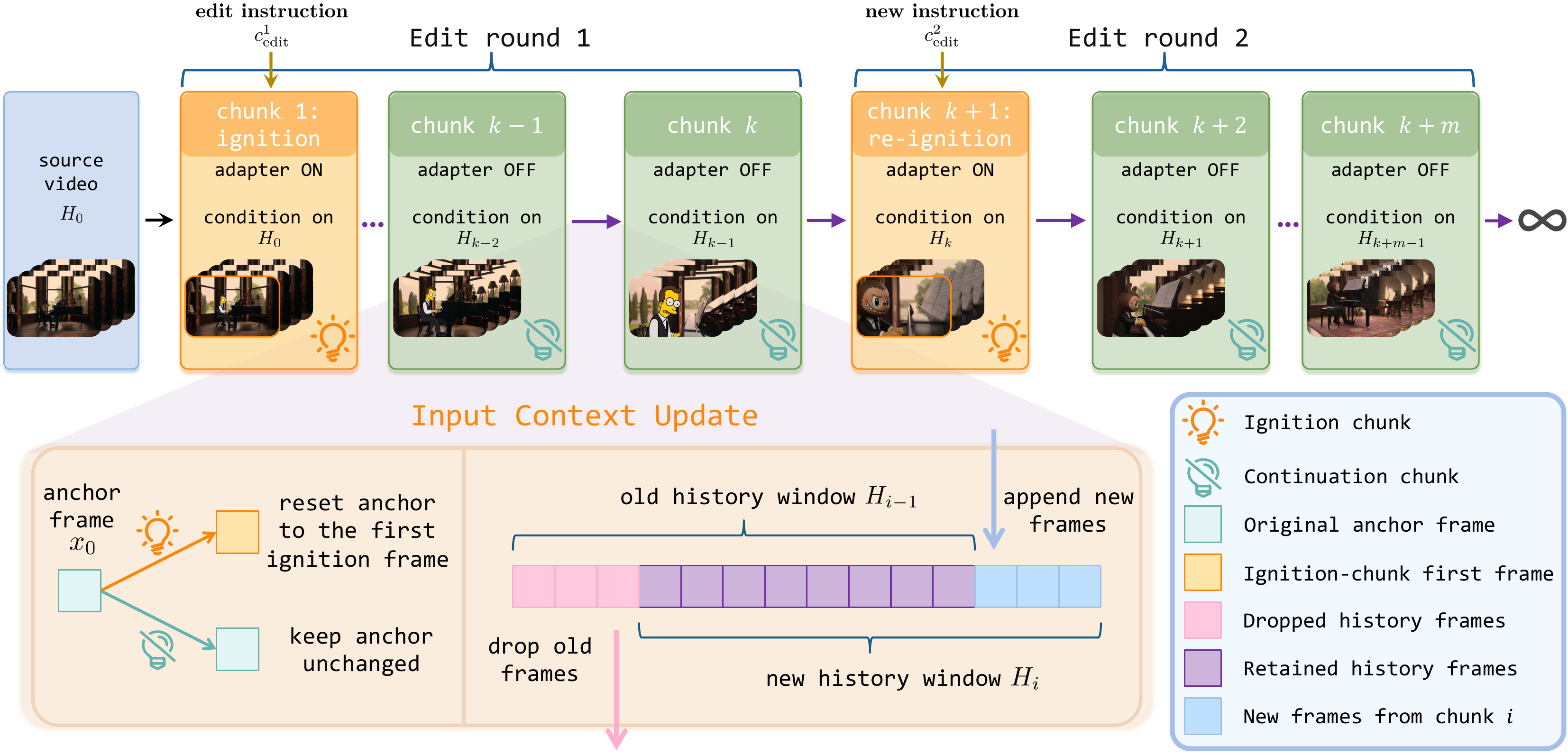}
    \caption{Inference pipeline of our method. Each edit round starts with an ignition chunk, after which subsequent chunks continue the stream.}
    \label{fig: method pipeline}
    \vspace{-10pt}
\end{figure}

In long video generation, our backbone model relies on a sliding history window and a fixed anchor frame $x_0$. The history window provides the direct condition for future continuation. The anchor frame $x_0$, together with the initial scene prompt, helps keep the generated stream from drifting. 
For repeated editing, we use these two components differently.
First, we keep the original sliding-window design. After each chunk, we append its frames and keep only the most recent window, so the memory cost stays constant as edits accumulate. 
Second, we update the anchor frame with the current edit. After the adapter generates the ignition chunk for an edit instruction, we reset $x_0$ to the first edited frame in the new chunk. This anchor is then held fixed during the following backbone continuations until the next edit instruction arrives. Resetting the anchor helps later chunks continue the edited stream rather than revert to the source. 
The window provides recent edited content, while the moving anchor gives a stable reference for the current edit. Together with the corruption-trained adapter, this design supports infinite edit rounds without unbounded memory or accumulated drift. At each ignition, we also replace the backbone's scene prompt with the edit instruction $c_{\text{edit}}$, so the following continuations stay aligned with the current edit. The update process for the history window and moving anchor frame is shown in Figure~\ref{fig: method pipeline}.

\section{Experiments}\label{sec:experiments}

\subsection{Experimental Settings}

\paragraph{Implementation Details.}
We use the 14B Helios-Distilled model~\cite{helios} as our backbone and train only the edit adapter. By default, edited videos are generated at  $384\times640$ resolution and $16$ fps. All training and evaluation are conducted on 32 NVIDIA H20 GPUs. The full training hyperparameters are provided in the Appendix.

\paragraph{Benchmark.}
No existing benchmark matches our setting. Established video-editing benchmarks such as OpenVE~\cite{OpenVE} focus on in-place editing. Their metrics rely on source-to-target frame correspondence. In our task, the target segment comes after the input and has no direct frame-level correspondence with it. These metrics are therefore not suitable for evaluation. 
To address it, we build an out-of-distribution (OOD) sequential-editing benchmark. Source clips are sampled from UltraVideo~\cite{ultravideo} and are disjoint from the training data. We use 200 source videos covering five entity categories, such as humans and scenery.
Each test sample contains a source video and its caption describing the source scene. Given this source, we use Gemini-3-Flash~\cite{team2023gemini} to generate a chain of three scene-grounded edit instructions. The instructions cover 15 edit types grouped into four categories: entity transformation, stylization, camera movement, and motion transfer. Each method under test then produces three temporally continuous segments, one for each instruction. Each segment spans four chunks, giving twelve chunks per sample. Since the instructions are grounded in each video's scene description, no evaluation instruction overlaps the training set. This format directly tests the sequential regime defined in Section~\ref{subsec: task description preliminary}.

\paragraph{Baselines.}
As discussed in Section~\ref{sec:related-work}, no prior method directly addresses infinite video editing. We therefore compare with five representative baselines from three families: \textit{Pure Backbone}, \textit{In-Place Editing}, and \textit{Prompt Switching}. The \textit{Pure Backbone} baseline uses the frozen Helios~\cite{helios} backbone. It replaces the source prompt with the edit instruction at each boundary and uses the same history input as our method. This baseline helps measure the gain from adding the edit adapter beyond prompt replacement alone. \textit{In-Place Editing} is represented by Lucy-Edit~\cite{decart2025lucyedit} and SANA-Streaming~\cite{zhao2026sana}. These methods rewrite the source clip while preserving frame-level correspondence. Comparing with them shows the difference between forward continuation and in-place rewriting. \textit{Prompt Switching} is represented by Anchor-Forcing~\cite{yang2026anchor} and Infinity-RoPE~\cite{yesiltepe2025infinity}. These methods continue an autoregressive stream after switching to a self-contained prompt, following a process similar to multi-shot video generation. They are architecturally close to our setting, but perform free generation rather than source-grounded editing. Because these methods do not take the source video as input, we mark them with $\dagger$ for distinction in the tables. All other source-conditioned methods, including pure backbone and in-place editing methods, receive the same source video for fairness.

\subsection{Quantitative Comparison}

We report quantitative results under two evaluation protocols: VBench metrics and a VLM-as-Judge study. We then analyze how each method behaves as edits accumulate.

\subsubsection{Evaluation with VBench Metrics}

\begin{table}[t]
\centering
\caption{Quantitative comparison on our sequential editing task (200 samples, 3 editing rounds each). All source-conditioned methods receive the same source video for fairness. Best results are in \textbf{bold}.}
\label{tab:main_comparison}
\resizebox{\linewidth}{!}{
\begin{tabular}{llccc|c}
\toprule
Category & Method & Camera Motion $\uparrow$ & Motion Smoothness $\uparrow$ & Temporal Flickering $\uparrow$ & Dynamic Degree \\
\midrule
Pure Backbone & Helios-Base (w/o Adapter)~\cite{helios} & 0.5494 & \textbf{0.9869} & 0.9641 & 0.6383 \\
\midrule
\multirow{2}{*}{In-Place Editing} & Lucy-Edit~\cite{decart2025lucyedit} & 0.5062 & 0.9808 & 0.9616 & 0.6067 \\
 & SANA-Streaming~\cite{zhao2026sana} & 0.5432 & 0.9858 & 0.9631 & 0.4750 \\
\midrule
\multirow{2}{*}{Prompt Switching} & Anchor-Forcing~\cite{yang2026anchor} & 0.4815 & 0.9775 & 0.9536 & 0.7917 \\
 & Infinity-RoPE~\cite{yesiltepe2025infinity} & 0.3580 & 0.9671 & 0.9344 & 0.8550 \\
\midrule
-- & Ours & \textbf{0.7654} & 0.9833 & \textbf{0.9660} & 0.7400 \\
\bottomrule
\end{tabular}
}
\end{table}

We first evaluate with a set of automatic, reference-free VBench~\cite{vbench} metrics. Following our task formulation, we focus on three evaluation aspects: edit-specific control, motion plausibility, and temporal stability. \emph{Camera Motion} measures the directional accuracy of camera edits with CoTracker2~\cite{karaev2024cotracker} point tracking. It is computed only on camera-type edits. \emph{Motion Smoothness} measures inter-frame plausibility using AMT~\cite{li2023amt} interpolation error. \emph{Temporal Flickering} is the mean absolute difference between adjacent frames. We report it on the final editing round to test stability after accumulated edits. We also report \emph{Dynamic Degree}, a RAFT~\cite{teed2020raft} flow-magnitude indicator, for reference only. It reflects the amount of motion in a clip, which depends on the content and edit type rather than edit fidelity.

As shown in Table~\ref{tab:main_comparison}, our method performs best on the main evaluation metrics. It achieves the best Camera Motion and Temporal Flickering, while staying close to the best Motion Smoothness. The gain in Camera Motion is the largest. Our method outperforms the second-best method by about $+0.22$, showing stronger control over directional camera edits. Temporal Flickering is measured on the third round, where accumulated degradation is most likely to appear. This result shows that our method maintains temporal coherence late in the editing chain.

\subsubsection{VLM-as-Judge Evaluation}

\begin{table}[t]
\centering
\caption{VLM-as-Judge evaluation on our infinite video editing task (3 editing rounds).
Best results are in \textbf{bold}.}
\label{tab:vlm_judge}
\resizebox{\linewidth}{!}{
\begin{tabular}{llcccc}
\toprule
Category & Method & Edit Faithfulness $\uparrow$ & Visual Quality $\uparrow$ & Preservation $\uparrow$ & Coherence $\uparrow$ \\
\midrule
Pure Backbone & Helios-Base (w/o Adapter)~\cite{helios} & 2.637 & 2.850 & 2.790 & 2.885 \\
\midrule
\multirow{2}{*}{In-Place Editing} & Lucy-Edit~\cite{decart2025lucyedit} & 2.863 & 2.723 & 2.665 & 2.560 \\
 & SANA-Streaming~\cite{zhao2026sana} & 3.303 & 3.093 & 3.060 & 3.030 \\
\midrule
\multirow{2}{*}{Prompt Switching} & Anchor-Forcing~\cite{yang2026anchor} & 2.545 & 2.620 & 1.830 & 3.480 \\
 & Infinity-RoPE~\cite{yesiltepe2025infinity} & 2.658 & 2.675 & 1.865 & 3.265 \\
\midrule
-- & \textbf{Ours} & \textbf{3.828} & \textbf{3.765} & \textbf{3.815} & \textbf{3.840} \\
\bottomrule
\end{tabular}
}
\end{table}

We also conduct a VLM-as-Judge evaluation with Gemini-3.5-Flash~\cite{team2023gemini} for a more comprehensive comparison. We score four dimensions on a 1--5 scale: \emph{Edit Faithfulness}, \emph{Visual Quality}, \emph{Scene Identity Preservation}, and \emph{Cross-Edit Coherence}. \emph{Edit Faithfulness} measures whether each segment follows its instruction. \emph{Visual Quality} measures the quality of the edited result. \emph{Scene Identity Preservation} checks whether the attributes that should remain unchanged are preserved. \emph{Cross-Edit Coherence} measures whether the three segments form a continuous sequence. Each sample is judged in two stages. The first call scores faithfulness and quality for each segment. The second call scores preservation and coherence over the full edit chain. 
Table~\ref{tab:vlm_judge} shows that our method ranks first on all four dimensions, with the largest margins in preservation and visual quality. The two baseline families show different weaknesses. \textit{In-Place Editors} re-render the given frames rather than continue the stream. As a result, they score lower on cross-edit coherence, since each round is rewritten in place instead of extending a single forward sequence. \textit{Prompt Switching} methods show the opposite pattern. They obtain relatively high coherence but fail on preservation, with scores below 2.0. This is because each segment is generated from a switched prompt without a source video for grounding. Our method is strong on both preservation and coherence, showing that its edits are grounded in the source content while still forming a continuous edited stream.

\subsubsection{Stability across Editing Rounds}

\begin{table}[t]
\centering
\caption{Edit faithfulness across sequential editing rounds. Our method maintains stable performance ($\Delta < 0.06$ between rounds), demonstrating robust sequential editing capability without degradation.}
\label{tab:vlm_per_round}
\resizebox{.5\linewidth}{!}{
\begin{tabular}{lccc|c}
\toprule
Method & Edit 1 & Edit 2 & Edit 3 & Std $\downarrow$ \\
\midrule
Helios-Base (w/o Adapter) ~\cite{helios}& 2.590 & 2.740 & 2.580 & 0.072 \\
Lucy-Edit~\cite{decart2025lucyedit} & 2.490 & 2.980 & 3.120 & 0.268 \\
SANA-Streaming~\cite{zhao2026sana} & 3.260 & 3.295 & 3.355 & 0.039 \\
Anchor-Forcing$^\dagger$~\cite{yang2026anchor} & 1.530 & 2.955 & 3.150 & 0.715 \\
Infinity-RoPE$^\dagger$~\cite{yesiltepe2025infinity} & 1.655 & 3.215 & 3.105 & 0.694 \\
\midrule
\textbf{Ours} & \textbf{3.860} & \textbf{3.805} & \textbf{3.820} & \textbf{0.023} \\
\bottomrule
\end{tabular}
}
\end{table}

\begin{wrapfigure}{r}{0.42\textwidth}
    \centering
    \vspace{-\intextsep}
    \includegraphics[width=0.42\textwidth]{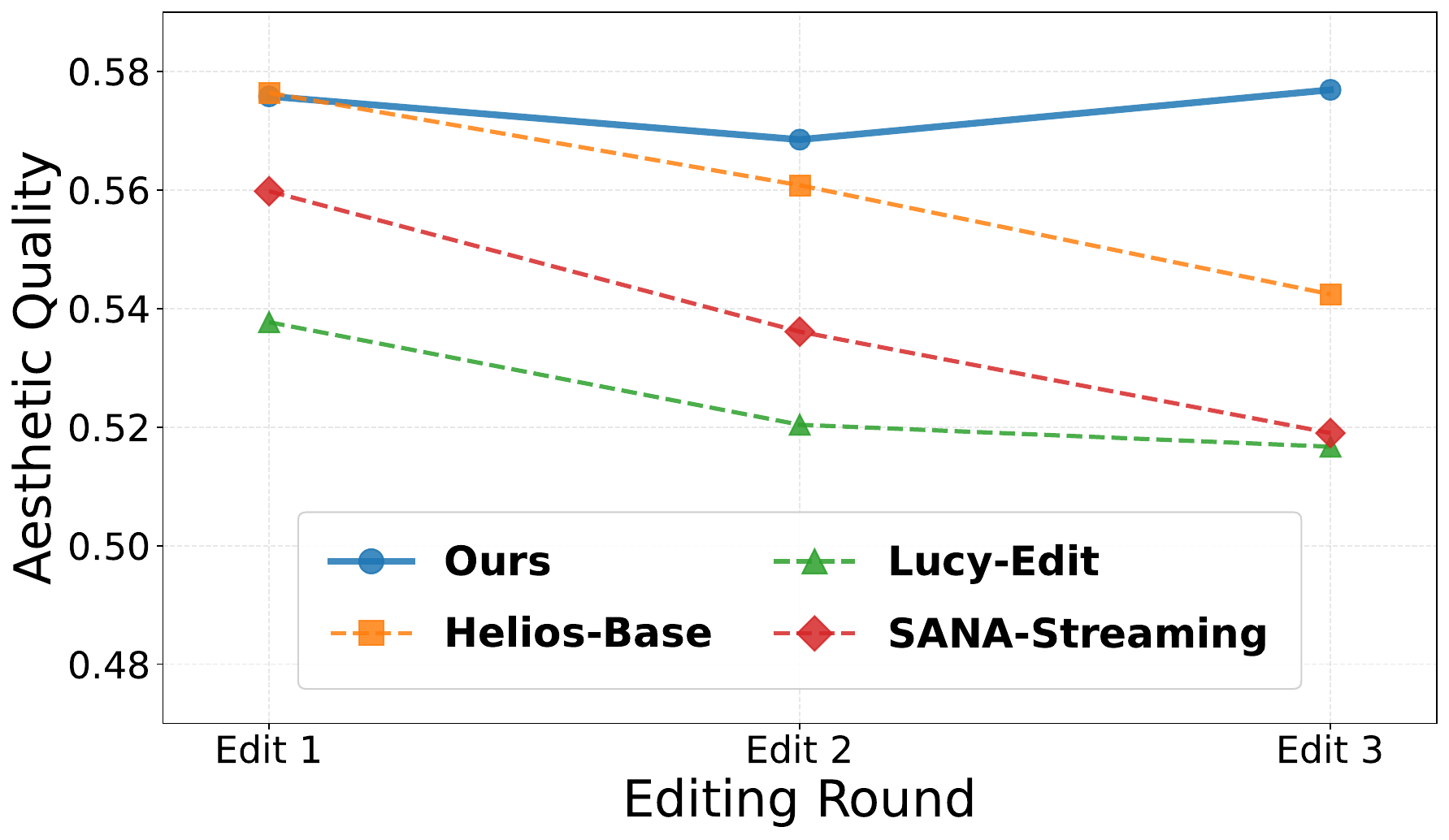}
    \vspace{-10pt}
    \caption{The aesthetic quality score recorded with different editing round.}
    \vspace{-10pt}
    
    \label{fig: aesthetic degradation}
\end{wrapfigure}

A key requirement of infinite editing is that quality and editing ability should not decay as edits accumulate. Figure~\ref{fig: aesthetic degradation} tracks \emph{Aesthetic Quality} computed by VBench across the three editing rounds. We restrict this comparison to source-conditioned methods and exclude the prompt switching family. Since prompt switching methods take no source video as input, their outputs are not anchored to the same visual content. Therefore, their aesthetic trajectory is not directly comparable. Among the source-conditioned methods, the baselines degrade over rounds and drop by 0.02 to 0.04 after the first round. In contrast, our method stays nearly flat ($\Delta \approx 0$), which confirms that it maintains visual quality under repeated editing.

Table~\ref{tab:vlm_per_round} reports edit faithfulness for each round. Our method ranks first at every step. Each baseline family shows a different weakness. The \textit{Pure Backbone} remains weak across all rounds and cannot handle diverse edits in the chain. The \textit{Prompt Switching} methods vary strongly across rounds. Their first segment has no source video to rely on and starts much lower than the others. Later segments improve only after the methods can build on their own generated stream. The \textit{In-Place Editors} appear to improve over rounds, but this trend is misleading. At each round, they re-edit their previous output. The content is therefore rewritten repeatedly, drifts from the source, and loses fidelity, as also shown in Figure~\ref{fig: aesthetic degradation}. This creates a looser canvas where later instructions are easier to impose and easier to score. To show round-level stability, we also report the standard deviation across rounds. Our method is the only one that remains both high and stable, staying near 3.8 from the first edit to the last (std\,$=$\,0.023). These results support our design choice: cross-attention conditions each step on both the current edit instruction and the recent history, which preserves editing ability across rounds.

\subsection{Qualitative Comparison}

\begin{figure}[t]
    \centering
    \includegraphics[width=\linewidth]{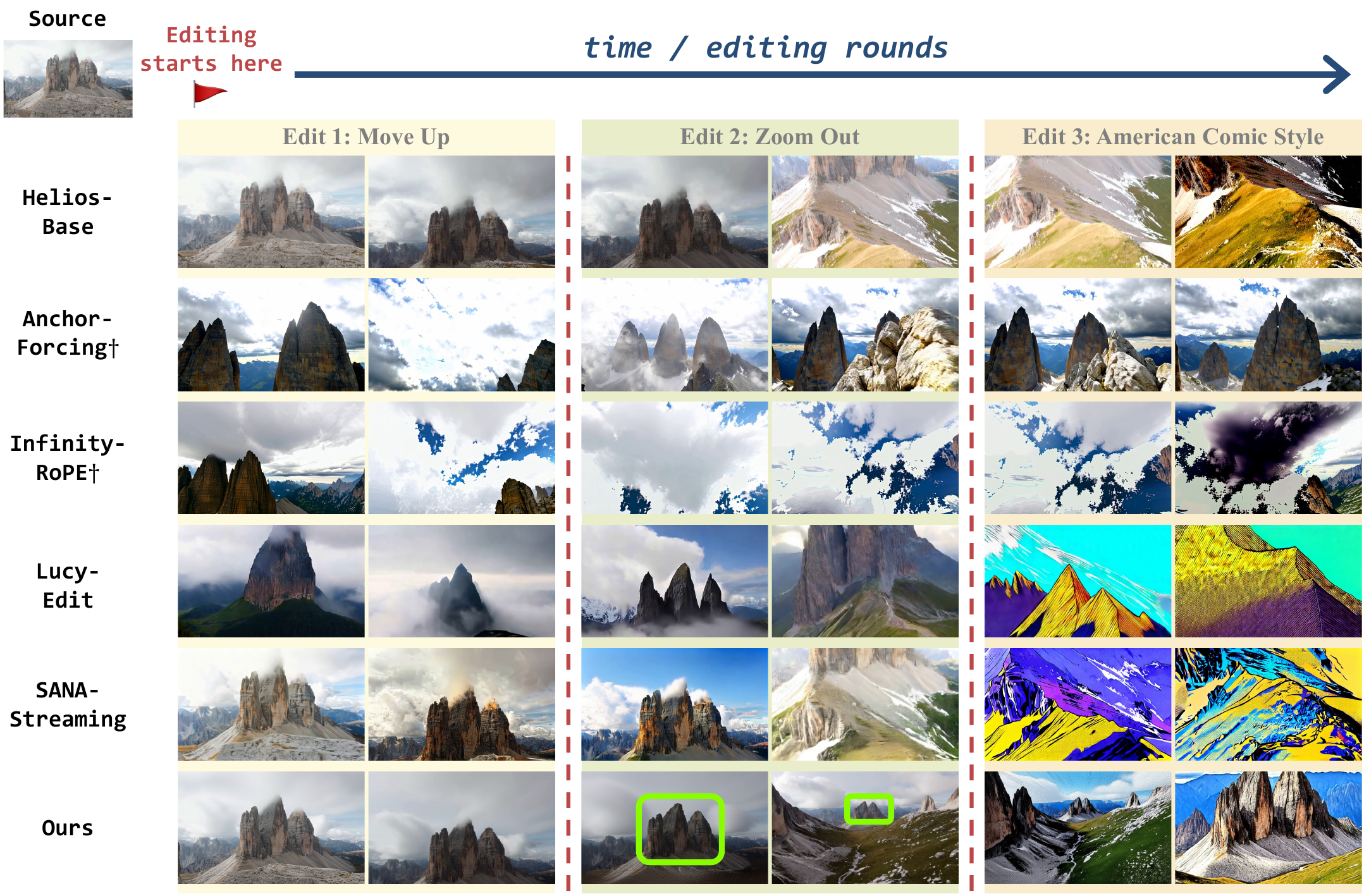}
    \caption{Qualitative comparison with various methods on the streaming video editing task, where a sequence of editing instructions (here camera-movement and style edits) is applied continuously to a scenery source video. All methods share the same source video and editing instructions. Methods marked with $\dagger$ take no source video as input, while all other conditions are kept identical for fairness.}
    \label{fig:comparison_case1}
\end{figure}

\begin{figure}[t]
    \centering
    \includegraphics[width=\linewidth]{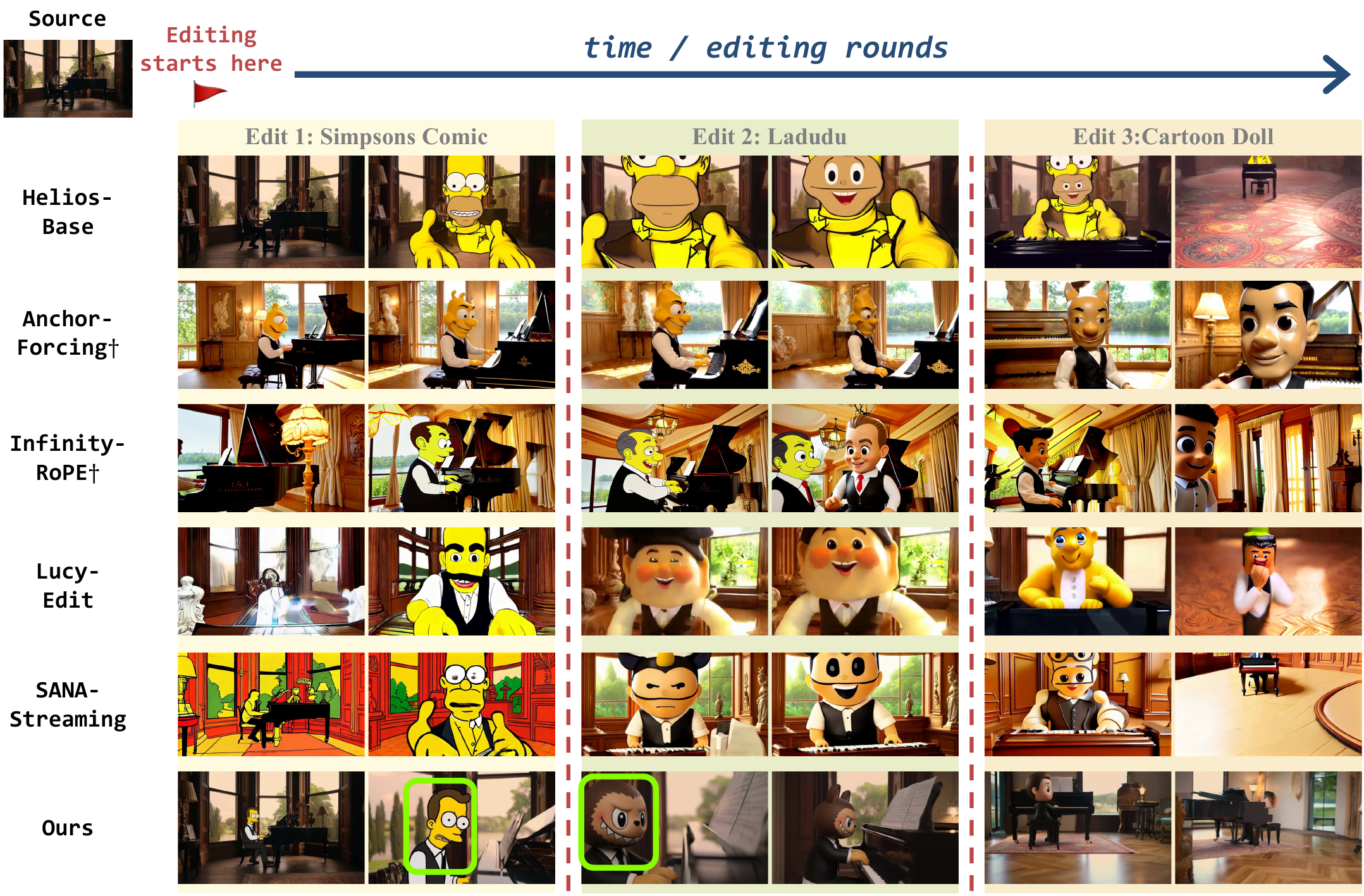}
    \caption{Qualitative comparison with various methods on the streaming video editing task, where a sequence of editing instructions (here entity-transformation edits) is applied continuously to a human-centric source video. All methods share the same source video and editing instructions. Methods marked with $\dagger$ take no source video as input, while all other conditions are kept identical for fairness.}
    \label{fig:comparison_case2}
\end{figure}

We present two qualitative cases from our benchmark. Figure~\ref{fig:comparison_case1} applies a chain of camera-movement and style edits to a scenery source. Figure~\ref{fig:comparison_case2} applies entity-transformation edits to a human-centric source. In both cases, each method receives the same source and the same three successive instructions. Each row shows the frames produced by one method as edits accumulate.

Our method applies each instruction at the intended point in the stream and carries the result forward. Within a segment, the edit takes effect at the first generated chunk and then propagates to the following chunks. This behavior comes from our inference regime in Section~\ref{subsubsec: history window and anchor resetting}, which keeps later chunks conditioned on recently edited content. In Figure~\ref{fig:comparison_case1}, the \textcolor[HTML]{89FC00}{green} boxes track the same mountain peak. Under the \emph{zoom-out} instruction, our method pulls the camera back so that the peak takes a smaller part of the frame. This shows that the edit is applied cleanly within a single segment. Across segment boundaries, the stream also stays continuous rather than restarting. In Figure~\ref{fig:comparison_case2}, the \textcolor[HTML]{89FC00}{green} boxes mark the last frame of the first segment and an early frame of the second segment. The character changes from a \textit{Simpsons-comic} style to \textit{Ladudu}, while the background remains continuous. The three edits therefore appear as one evolving video rather than three separate clips.

The competing methods fall short in different ways. Some fail to realize the intended edit. Others apply the edit but drift from the source subject or scene, or break continuity across segments. Our method is the only one that keeps the edit faithful, preserves the source content, and maintains a continuous stream through all three rounds. This is consistent with its leading scores on preservation and cross-edit coherence in Table~\ref{tab:vlm_judge}.

\subsection{Long-Video Editing}

\begin{figure}[t]
    \centering
    \includegraphics[width=\linewidth]{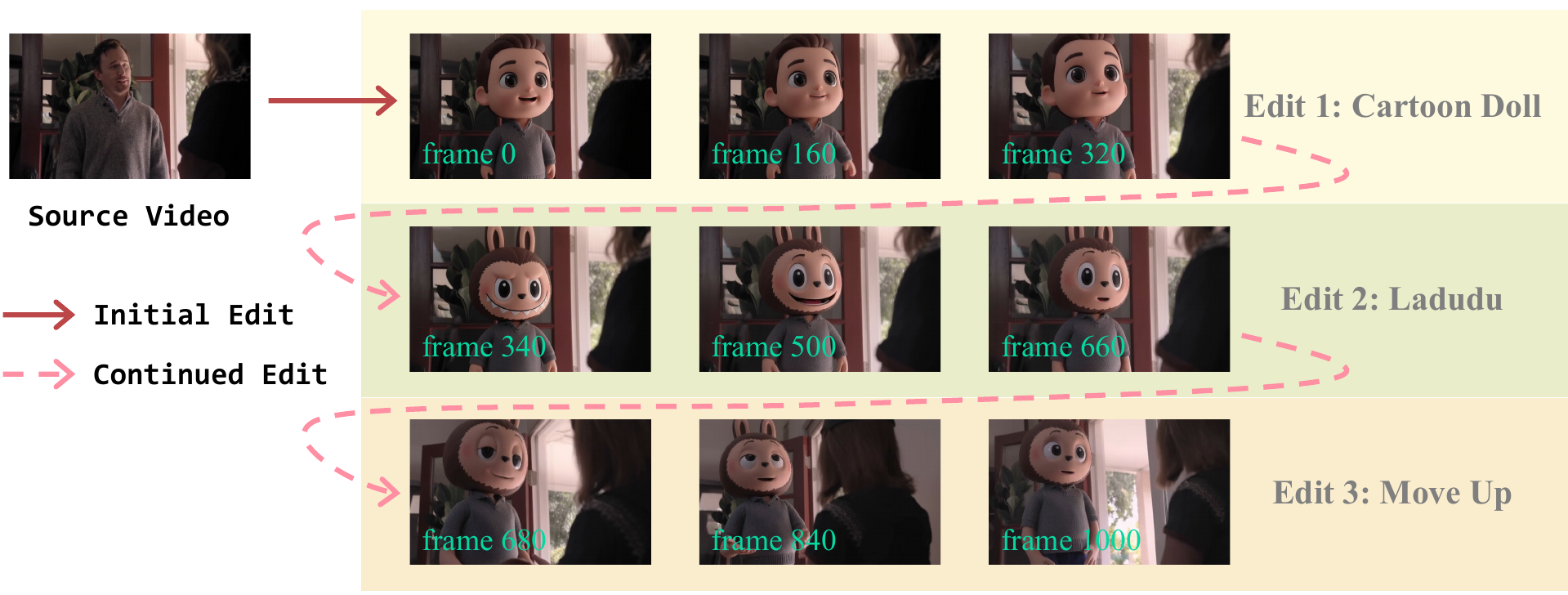}
    \caption{Long-video streaming editing. We lengthen each editing segment so that the full sequence exceeds 1000 frames. Our method stays visually stable and keeps realizing each instruction. The edited attributes also persist across segments.}
    \label{fig:long_video_case1}
\end{figure}

We further test the infinite-editing regime by lengthening each editing segment, so the full sequence exceeds 1000 frames, as shown in Figure~\ref{fig:long_video_case1}. Our method remains visually stable over this longer horizon and continues to follow each instruction without quality collapse. This stability comes from the Helios backbone, whose streaming generation prior is kept frozen. The adapter adds edit control on top of this prior without weakening its ability to sustain long videos. Edited attributes also persist across segments. For example, while the third instruction (\textit{move up}) is applied, the Ladudu appearance introduced in the second segment is still preserved. This shows that both editing ability and content memory remain stable under long-horizon generation. More long-video cases are provided in the Appendix.
\section{Conclusion}

In this paper, we introduced \textbf{infinite video editing}, a task where edits are applied to an ongoing video stream beyond a fixed input clip. Unlike in-place editing, the model must generate the next segment while applying the requested edit. This setting requires faithful continuation and stable quality as edits accumulate. To address these challenges, we proposed \textbf{InfinityEdit}, a lightweight edit adapter for a frozen streaming video generator. The adapter has three modules: history cross-attention for grounding the new segment in the input history, temporal causal self-attention for forward temporal propagation, and edit cross-attention for injecting the edit instruction. At inference, the adapter is activated only when an edit request arrives, while the frozen generator carries the edited stream forward with a sliding history window and a reset anchor frame. 
Experiments show that InfinityEdit outperforms baseline methods. It follows edit requests more faithfully, continues the stream without falling back to frame-wise rewriting, and remains stable as edits accumulate over long sequences.

\section{Future Work}

Our editor currently uses only natural-language instructions. Adding image or video examples as references could give users more control over the target appearance, especially for edits that are hard to describe in words. A second direction is the transition at edit boundaries. Although the continuation remains coherent, the switch to specific-type instructions can still be abrupt. Smoother transitions between successive edits remain an important direction for future work.

\clearpage

\bibliographystyle{plainnat}
\setlength{\bibhang}{0pt}
\setlength\bibindent{0pt}
\bibliography{main}

\end{document}